\documentclass[letterpaper, 10 pt, conference]{ieeeconf}  

\IEEEoverridecommandlockouts                              

\title{\LARGE \bf
Dashing for the Golden Snitch: Multi-Drone Time-Optimal Motion Planning with Multi-Agent Reinforcement Learning
}
\usepackage{graphicx}
\usepackage{subfigure}
\usepackage{amsmath}
\usepackage{mathtools}
\usepackage{float}
\usepackage{amsfonts,amssymb}
\usepackage{lipsum}
\usepackage{hyperref}
\hypersetup{hidelinks,
	colorlinks=true,
	allcolors=blue,
	pdfstartview=Fit,
	breaklinks=true}
\usepackage{algorithm}
\usepackage{algorithmicx}
\usepackage{algpseudocode}
\usepackage{makecell}
\usepackage{multirow}
\usepackage{cite}
\usepackage{color}
\usepackage[all]{hypcap}

\author{Xian Wang$^{1}$, Jin Zhou$^{1}$, Yuanli Feng$^{1}$, Jiahao Mei$^{2}$, Jiming Chen$^{1}$, and Shuo Li$^{1}$
\thanks{$^{1}$Authors are with the College of Control Science and Engineering, Zhejiang University, Hangzhou 310027, China.
        {\tt\small shuo.li@zju.edu.cn}
        }%
\thanks{$^{2}$Jiahao Mei is with the Department of Automation, Zhejiang University of Technology, Hangzhou 310023, China.
        }%
\thanks{This work was supported in part by NSFC under Grants 62203385, 62088101.}
}

\begin{document}

\maketitle
\thispagestyle{empty}
\pagestyle{empty}


\begin{abstract}
Recent innovations in autonomous drones have facilitated time-optimal flight in single-drone configurations, and enhanced maneuverability in multi-drone systems by applying optimal control and learning-based methods. However, few studies have achieved time-optimal motion planning for multi-drone systems, particularly during highly agile maneuvers or in dynamic scenarios. This paper presents a decentralized policy network using multi-agent reinforcement learning for time-optimal multi-drone flight. To strike a balance between flight efficiency and collision avoidance, we introduce a soft collision-free mechanism inspired by optimization-based methods. By customizing PPO in a \textit{centralized training, decentralized execution} (CTDE) fashion, we unlock higher efficiency and stability in training while ensuring lightweight implementation. Extensive simulations show that, despite slight performance trade-offs compared to single-drone systems, our multi-drone approach maintains near-time-optimal performance with a low collision rate. Real-world experiments validate our method, with two quadrotors using the same network as in simulation achieving a maximum speed of {\normalfont$13.65\ \text{m/s}$} and a maximum body rate of {\normalfont$13.4\ \text{rad/s}$} in a {\normalfont$5.5\ \text{m} \times 5.5\ \text{m} \times 2.0\ \text{m}$} space across various tracks, relying entirely on onboard computation [\href{https://youtu.be/KACuFMtGGpo}{\textbf{\emph{video}}}\footnote[3]{https://youtu.be/KACuFMtGGpo}][\href{https://github.com/KafuuChikai/Dashing-for-the-Golden-Snitch-Multi-Drone-RL}{\textbf{\emph{code}}}\footnote[4]{https://github.com/KafuuChikai/Dashing-for-the-Golden-Snitch-Multi-Drone-RL}].
\end{abstract}

\hypersetup{hidelinks,
	colorlinks=true,
	allcolors=black,
	pdfstartview=Fit,
	breaklinks=true}

\section{INTRODUCTION}
Quadrotors have shown significant potential for time-critical tasks such as delivery, inspection, and search-and-rescue operations, owing to their unmatched agility among unmanned aerial vehicles. Extensive studies have contributed to single-drone time-optimal trajectory generation through waypoints \cite{bry2015aggressive, foehn2021time, wang2022geometrically, fork2023euclidean, qin2023time}. These optimization-driven approaches preclude real-time applications due to computational intensity. Alternatively, those that utilize polynomial expressions are often constrained by limited order or segment numbers, making it challenging to fully harness the potential of drones. To facilitate online optimal control, Model Predictive Contouring Control (MPCC) \cite{romero2022model,romero2022time} enables tracking predefined trajectories while performing time-optimal online replanning. 
\begin{figure}[htbp]
    \centering
    \includegraphics[width=0.43\textwidth, trim={0.4cm 0.3cm 0.2cm 0.5cm}, clip]{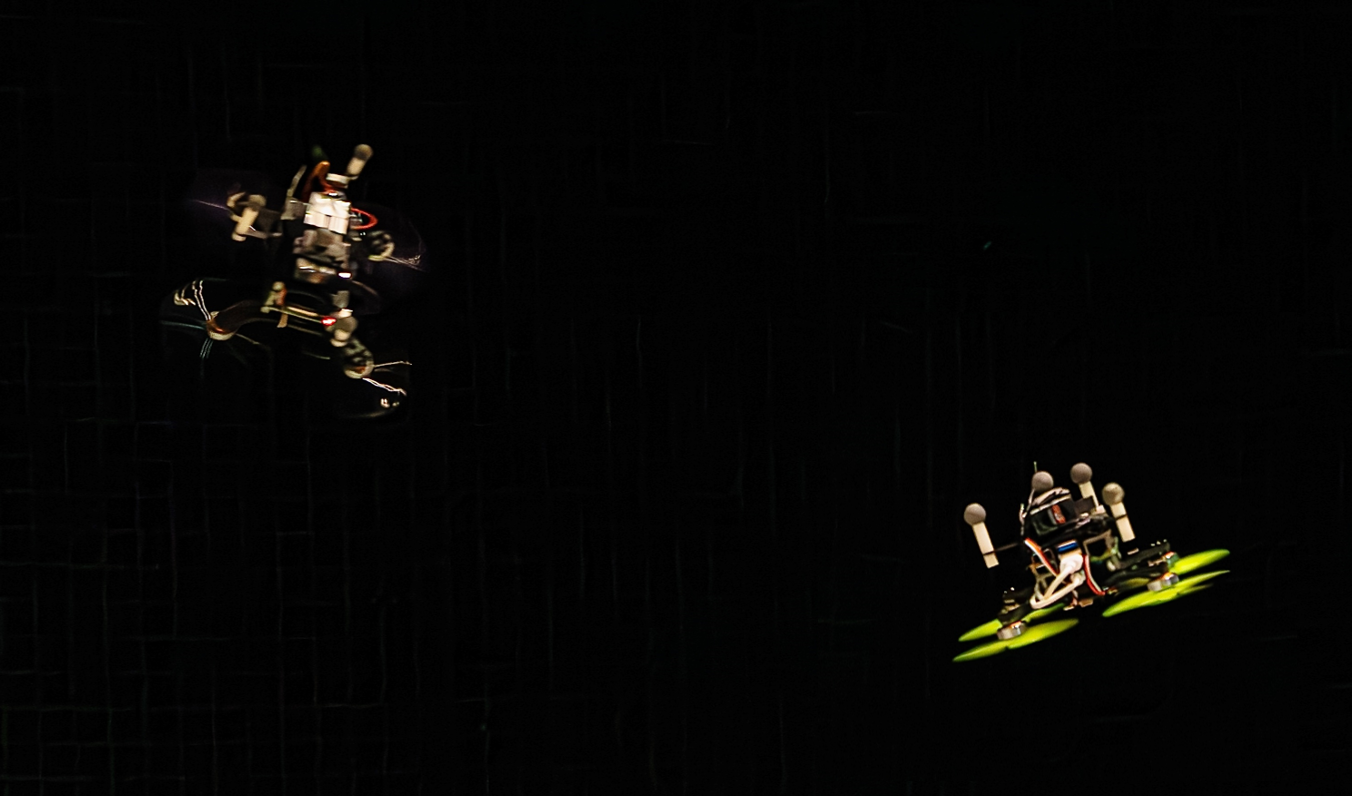} 
    \vspace{-0.16cm}
    \caption{Two quadrotors executing highly agile maneuvers, guided by onboard policy networks in a real-world flight.}
    \vspace{-0.63cm}
    \label{fig:real_platform}
\end{figure}
The success of reinforcement learning (RL) in enhancing single-drone agility has recently reached remarkable milestones, with PPO-based frameworks achieving near-time-optimal flight and champion-level racing performance \cite{Autonomous2021song,song2023reaching,kaufmann2023champion,ferede2024end}. These advances demonstrate RL's capacity to handle aggressive maneuvers under uncertainty, even when drones rely solely on onboard sensor inputs. However, aerial tasks can be executed more efficiently with quadrotor swarms. Real-world deployment of fully autonomous multi-drone systems faces challenges such as limited sensing ranges, unreliable communication, and restricted computational resources \cite{guo2024mighty}. Despite these challenges, researchers strive to optimize multi-drone agility and cooperation.

In multi-drone systems, a key challenge is to plan trajectories for each drone while avoiding inter-drone collisions—a problem nonexistent in single-agent cases. As the size of the swarm increases, the computational demands of centralized planning grow exponentially. Therefore, decentralized trajectory planning is becoming essential to address this issue efficiently. Recently, several comprehensive studies have been conducted that focus on drone swarms, such as EGO-Swarm \cite{zhou2022swarm}, MADER \cite{tordesillas2021mader}, DLSC \cite{park2023dlsc}, and AMSwarm \cite{adajania2023amswarm, adajania2023amswarmx}. These studies tackle various aspects of trajectory optimization, improving swarm navigation in both simulations and real-world environments. However, most research has concentrated on partial dynamic models, which limits the drones' agility. Shen et al. \cite{shen2023aggressive} employ the complementary progress constraint \cite{foehn2021time} in multi-drone scenarios to generate time-optimal trajectories for racing swarms navigating predefined waypoints. Nevertheless, this method requires hours of calculation, preventing online applications. Overall, significant potential remains for future research to enhance the aggressiveness of drone swarms in more practical scenarios.

In contrast to traditional methods, an increasing number of studies are adopting RL-based approaches, leveraging PPO \cite{PPO} combined with the \textit{centralized training, decentralized execution} (CTDE) extension \cite{NEURIPS_MAPPO}, to learn decentralized policies directly in simulation environments. Specifically, RL-based solutions have been applied to obstacle avoidance \cite{bhattacharya2024vision} and multi-drone racing tasks at speeds up to $2\ \text{m/s}$ \cite{kang2024autonomous}, with recent work \cite{zhang2024back} demonstrating communication-free multi-agent navigation via vision-based policies. However, existing methods either treat flight time as a secondary objective or lack experimental validation of aggressive maneuvers in real-world settings, thereby leaving time-optimal flight with collision avoidance unresolved for agile multi-drone systems.

In this paper, we focus on leveraging multi-agent reinforcement learning to train decentralized policy networks. This approach enables multiple drones to traverse waypoints with near-time-optimal performance while avoiding inter-drone collisions. Additionally, the networks are computationally lightweight, enabling high-frequency inference solely with onboard resources.

Hence, in this study, we 

\begin{enumerate}
    \item Develop a decentralized policy network for multi-drone time-optimal motion planning using multi-agent reinforcement learning. Inspired by soft constraints in optimization-based methods, we propose a soft collision-free mechanism incorporating continuous penalties and safety margins, enabling collision-aware learning without compromising time-optimality.
    \item Enhance learning efficiency through a modified PPO algorithm integrated with invalid-experience masking, value normalization, and a \textit{centralized training, decentralized execution} (CTDE) framework.
    \item Demonstrate competitive performance through simulations and real-world experiments. Our policies achieve competitive performance with low collision rates in thousands of simulations. In physical deployments, using the same networks as in simulation, two quadrotors achieve a maximum speed of $13.65\ \text{m/s}$ and a peak body rate of $13.4\ \text{rad/s}$ in a $5.5\ \text{m} \times 5.5\ \text{m} \times 2.0\ \text{m}$ space, relying entirely on onboard computation.
\end{enumerate}

\section{METHODOLOGY}
\begin{figure*}[htbp]
    \centering
    \includegraphics[width = 1.0\textwidth, trim={60pt 88pt 60pt 76pt}, clip]{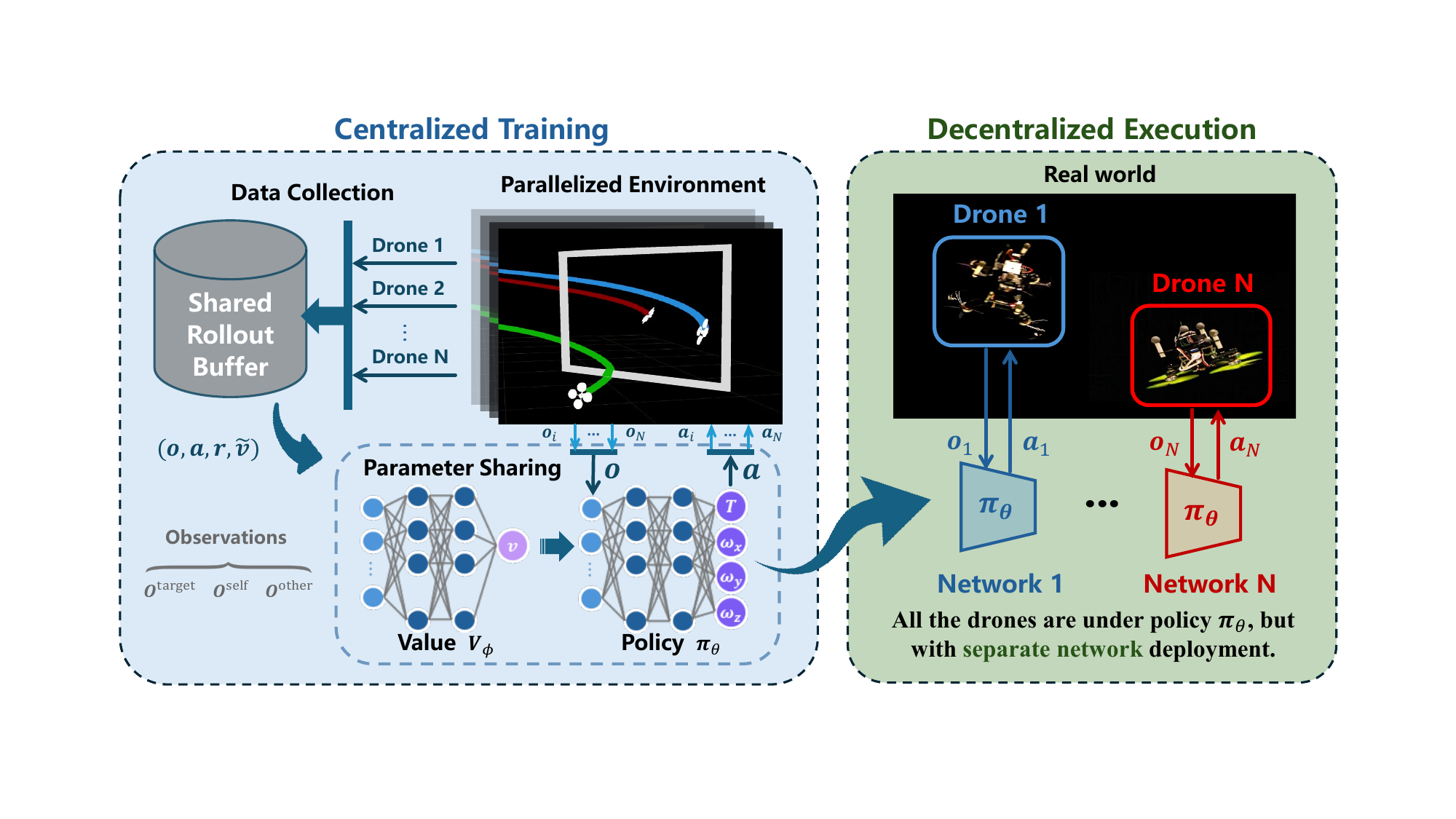} 
    \caption{Overview of the proposed method illustrating CTDE framework. During the training, all drones share a common policy network $\pi_\theta$ and a value network $V_\phi$, with data collected in parallel and stored in a shared rollout buffer. For deployment, each drone independently executes the policy in a decentralized manner, using identical network parameters while making decisions based on its local information.} 
    \label{fig:proposed_sketch}
    \vspace{-0.4cm}
\end{figure*}
\subsection{The multi-agent RL problem}
This paper focuses on enabling each drone to navigate through waypoints using local information while avoiding inter-drone collisions via reinforcement learning (RL). To achieve this, we frame this problem as a Decentralized Partially Observable Markov Decision Process (DEC-POMDP). 

We define the DEC-POMDP of an $N$-drone scenario by a 6-tuple $({\mathcal{S}}, {\mathcal{A}}, {\mathcal{O}}, {\mathcal{R}}, {\mathcal{P}}, {\gamma})$. The state space ${\mathcal{S}}$ represents all possible global states $\mathbf{s}$ (where $\mathbf{s} \in {\mathcal{S}}$), including the positions, velocities, and orientations of all drones, along with relevant external factors such as waypoints and historical data. Throughout an episode, at the timestep $t$, drone $i$ selects actions $\mathbf{a}_i^t$ from a shared action space ${\mathcal{A}}$, informed by its own local observations $\mathbf{o}_i^t = \mathcal{O}(\mathbf{s}^t;i)$ from the global state ${\mathbf{s}^t}$. The state transition probability ${\mathcal{P}}(\mathbf{s}'|\mathbf{s},\mathbf{a})$ dictates transitions based on the joint actions $\mathbf{a}=(\mathbf{a}_1,\dots,\mathbf{a}_N)$ taken by all the drones. The state transitions are governed by a simplified quadrotor dynamic model, discretized over time intervals $\Delta t$, as detailed in Section~\ref{sec:dynamics}.

The reward function $r_i^t = \mathcal{R}(\mathbf{s}^t, \mathbf{a}_i^t)$ encourages actions that balance the time-optimal flight with collision avoidance. The discount factor $\gamma \in [0, 1]$ ensures drones prioritize long-term efficiency over immediate rewards. The goal is to find an optimal policy $\pi_{\theta}$ that maximizes the cumulative rewards. For an $N$-drone system, this objective is formulated as:
\begin{equation}
    \begin{array}{*{2}{l}}
        \underset{\pi_{\theta}}{\text{argmin}} & {\mathbb{E}\left[\dfrac{1}{N}{\displaystyle \sum_{i=1}^{N}}{t_i}(\pi_{\theta})\right]}, \\[15pt]
        {\text{subject to}} & {\left\| \mathbf{p}_{i}^{t}-\mathbf{p}_{j}^{t} \right\|_2 \geq 2R,\ \forall i \neq j,\ \forall t \in [0,T],}
    \end{array}
    \label{equ:problem_description}
\end{equation}
where $T$ is the episode horizon, $t_i$ is the travel time for drone $i$ under policy $\pi_{\theta}$, $\mathbf{p}_i^t \in \mathbb{R}^3$ is the position of drone $i$ at timestamp $t$ and $R$ is the drone radius. The constraint enforces pairwise collision avoidance throughout the entire trajectory. Section~\ref{sec:obs-and-action} subsequently details the observation space, the action space, and the reward design.

\subsubsection{\textbf{Observations and actions}} \label{sec:obs-and-action}
Each drone's observation vector $\mathbf{o}_i$ comprises three components: \textit{waypoints} $\mathbf{o}_i^{\text{wp}}$, \textit{ego-state} $\mathbf{o}_i^{\text{ego}}$, and \textit{neighbors} $\mathbf{o}_i^{\text{neigh}}$. The waypoints observation is defined as $\mathbf{o}_i^{\text{wp}} = [\Delta\mathbf{p}_{i_1}, \dots, \Delta\mathbf{p}_{i_k}, \dots] \in \mathbb{R}^{3W}, k \in \{1, \dots, W\}$, where $W=2$ specifies the number of future waypoints considered in the lookahead window. The relative position $\Delta\mathbf{p}_{i_k}$ is computed as:
\begin{itemize}
    \item \textit{Drone to target} ($k = 1$): The drone's relative position to the next immediate waypoint.
    \item \textit{Waypoint chain} ($k \geq 2$): The relative position between consecutive waypoints within the lookahead window.
\end{itemize}
The ego-state $\mathbf{o}_i^{\text{ego}} = [\mathbf{v}_i, \operatorname{vec}{\left(\mathbf{R}(\mathbf{q}_i)\right)}] \in \mathbb{R}^{12}$ includes velocity $\mathbf{v}_i$ and flattened rotation matrix $\operatorname{vec}{\left(\mathbf{R}(\mathbf{q}_i)\right)}$ derived from quaternion $\mathbf{q}_i$. The neighboring observations $\mathbf{o}_i^{\text{neigh}} = \bigoplus_{j \neq i} [\Delta\mathbf{p}_{ij}, \Delta\mathbf{v}_{ij}, \|\Delta\mathbf{p}_{ij}\|_2] \in \mathbb{R}^{7(N-1)}$ (where $\bigoplus$ denotes concatenation) encode relative kinematics between drones, where $\Delta\mathbf{p}_{ij} = \mathbf{p}_j - \mathbf{p}_i$ and $\Delta\mathbf{v}_{ij} = \mathbf{v}_j - \mathbf{v}_i$.

All observations are normalized before network processing: the relative positions of waypoints $\Delta\mathbf{p}_{i_k} \gets \Delta\mathbf{p}_{i_k} / k_p$, the ego-velocity $\mathbf{v}_i \gets \mathbf{v}_i / k_v$ and the neighboring features $\Delta\mathbf{p}_{ij} \gets \Delta\mathbf{p}_{ij}/k_{rp},\ \Delta\mathbf{v}_{ij} \gets \Delta\mathbf{v}_{ij}/k_{rv},\ \|\Delta\mathbf{p}_{ij}\|_2 \gets \|\Delta\mathbf{p}_{ij}\|_2/k_d$, where $k_p$, $k_v$, $k_{rp}$, $k_{rv}$, and $k_d$ are normalization constants listed in Table~\ref{tab:Simulation_Parameters}. The policy $\pi_\theta$ outputs normalized actions $\mathbf{a}_i = [\tilde{T}_{\text{cmd}}, \tilde{\boldsymbol{\omega}}_{\text{cmd}}]$, which are converted to physical commands via: 
\begin{equation}
T_{\text{cmd}} = \frac{\tilde{T}_{\text{cmd}} + 1}{2}T_{\text{max}}, \quad
\boldsymbol{\omega}_{\text{cmd}} = \tilde{\boldsymbol{\omega}}_{\text{cmd}} \boldsymbol{\omega}_{\text{max}},
\end{equation}
where $T_{\text{max}}$ and $\boldsymbol{\omega}_{\text{max}}$ represent the thrust-to-weight ratio and maximum body rates, respectively.

\subsubsection{\textbf{Rewards}}
The reward function guides the drones to attain the time-optimal objective. For each drone $i$, the per-timestep reward at timestep $t$ is calculated individually as:
\begin{align}
    {r}_{i}^{t} = {r}_{i}^{t,\text{target}} + {r}_{i}^{t,\text{smooth}} + {r}_{i}^{t,\text{safe}} + {r}_{i}^{t,\text{crash}},
\end{align}
consisting of four components. The target reward ${r}_{i}^{t,\text{target}}$ incentivizes progress toward the next waypoint $\mathbf{g}_i$, avoiding direct waypoint-reaching rewards to prevent risky trajectories. To achieve this, we replace direct rewards for reaching the waypoint with a virtual tangent sphere of radius $\eta d_\text{w}$ ($\eta=0.75$), smaller than the actual waypoint radius $d_\text{w}$:
\begin{align}
    r_i^{t,\text{target}} &= 
    \begin{cases}
        r_\text{arrival}, & \text{if}\ \|\mathbf{p}_i^t - \mathbf{g}_i\| < d_\text{w} \\
        L_i^{t-1} - L_i^t, & \text{otherwise}
    \end{cases} \\
    L_i^t &= \sqrt{\|\mathbf{p}_i^t - \mathbf{g}_i\|^2 - \left(\eta d_\text{w}\right)^2}, \quad 0<\eta<1.
\end{align}
This design promotes safe approach paths while maintaining time efficiency.

To penalize dynamically infeasible commands, the smoothness reward is defined as:
\begin{align}
    {r}_{i}^{t,\text{smooth}} = -\lambda_{1}{\left \| \boldsymbol{\omega}_{i}^{t} \right \|} - \lambda_{2} {\left \| \mathbf{a}_{i}^{t-1} - \mathbf{a}_{i}^{t} \right \|},
\end{align}
where $\boldsymbol{\omega}_{i}^{t}$ is the body rate of drone $i$, and $\mathbf{a}_{i}^{t-1}$, $\mathbf{a}_{i}^{t}$ denote consecutive actions. The second term penalizes abrupt control variations.

For collision avoidance, the safe reward ${r}_{i}^{t,\text{safe}}$ imposes expensive costs when the drones are either close to or approaching each other at high relative speeds. Discarding the timestep $t$ notation for simplicity, we arrive at the following expression:
\begin{align}
    r_i^{\text{safe}} &= \sum_{j \neq i} c_{ij} \left( \lambda_3 r_{ij}^{\text{dist}} + \lambda_4 \|\Delta\mathbf{v}_{ij}\| r_{ij}^{\text{vel}} \right),
\end{align}
where the proximity penalty $r_{ij}^{\text{dist}}$ and velocity-dependent penalty $r_{ij}^{\text{vel}}$ are:
\begin{align}
    r_{ij}^{\text{dist}} &= 
    \begin{cases}
        \min\left\{ e^{-\beta (\|\Delta\mathbf{p}_{ij}\| - 2R)}, 1 \right\}, & c_{ij} < 0 \\
        0, & \text{otherwise}
    \end{cases} \\
    r_{ij}^{\text{vel}} &= 
    \begin{cases}
        \left[ \mathrm{clip}\left( 1 - \frac{\|\Delta\mathbf{p}_{ij}\| - 3R}{d_\text{w}}, 0, 1 \right) \right]^2, & c_{ij} < 0 \\
        0, & \text{otherwise}
    \end{cases}
\end{align}
Here, $\Delta\mathbf{p}_{ij} = \mathbf{p}_j - \mathbf{p}_i$ and $\Delta\mathbf{v}_{ij} = \mathbf{v}_j - \mathbf{v}_i$ denote relative position and velocity vectors between the $i$-th and $j$-th drone. The factor $c_{ij} = \cos\angle(\Delta\mathbf{p}_{ij}, \Delta\mathbf{v}_{ij})$ evaluates collision risk directionality. The $\mathrm{clip}(x,0,1)$ function ensures $x \in [0,1]$. The decay rate $\beta=15$ is set to balance penalty sensitivity.

A crash reward ${r}_{i}^{t,\text{crash}}$ penalizes collisions and boundary violations. Unlike episodic termination strategies, we apply a soft collision-free mechanism inspired by optimization-based soft constraints, allowing temporary constraint violations in Equation~\ref{equ:problem_description} through continuous penalties. This enables simultaneous learning of collision avoidance and primary objectives.

We introduce a safety tolerance $\tau$ to relax collision constraints. Boundary violations trigger immediate episode termination, while collisions within $2R+\tau$ incur linearly cumulating penalties per incident:
\begin{align}
    r_i^{t,\text{crash}} &= 
    \begin{cases}
        -r_\text{collision}, & \exists j \neq i: \|\Delta\mathbf{p}_{ij}\| \leq 2R + \tau \\
        -r_\text{boundary}, & \mathbf{p}_i^t \notin \mathcal{W} \\
        0, & \text{otherwise}
    \end{cases}
\end{align}
where $\mathcal{W}$ defines the valid workspace, and $\tau=R$ provides 50\% safety margin beyond the nominal collision radius $2R$.

\subsection{Policy training}
\subsubsection{\textbf{Multi-agent RL algorithm}}
Building on decentralized multi-agent frameworks \cite{ICRA_multi_robot, de2020independent, wang2024end}, we adopt Independent Proximal Policy Optimization (IPPO) for its robustness in heterogeneous multi-agent tasks without domain-specific adaptations. To improve convergence and training stability, we implement two enhancements: (1) \textit{invalid-experience masking} to filter data from crashed drones, and (2) \textit{value normalization} for the critic network $V_\phi$ \cite{NEURIPS_MAPPO}. Leveraging agent homogeneity, we employ parameter sharing across all drones \cite{param_share_mappo5, param_share_mappo33}.

Drawing from the CTDE paradigm, we train a shared policy network $\pi_\theta$ and centralized critic $V_\phi$ that estimates state values from the state space $\mathcal{S}$, as formalized in Algorithm~\ref{algo:IPPO}. 

The training loop operates as follows: At each timestep $t$, drone $i$ selects action $\mathbf{a}_i^t \sim \pi_\theta(\cdot|\mathbf{o}_i^t)$ based on its local observation $\mathbf{o}_i^t$. For any variable $x^t$ related to a single drone, we define $\mathbf{x}^t = [x_1^t, x_2^t, \dots, x_N^t]$ as the collection of that variable across all $N$ drones. After executing joint actions $\mathbf{a}^t$, all experiences are stored in a shared data buffer $D$, including rewards $\mathbf{r}^t$, normalized values $\tilde{\mathbf{v}}^t$, and crash masks $\mathbf{m}^t$ along with the corresponding observations and actions. During policy updates, we:
\begin{enumerate}
    \item Denormalize values using $\tilde{\mathbf{v}}^t$ and compute advantages via Generalized Advantage Estimation (GAE) \cite{GAE},
    \item Apply crash masks $\mathbf{m}^t$ to exclude invalid transitions,
    \item Optimize $\pi_\theta$ and $V_\phi$ using the PPO objective \cite{PPO}.
\end{enumerate}
Figure~\ref{fig:proposed_sketch} summarizes the full training-deployment workflow.

\begin{algorithm}
    \caption{Masked IPPO with Value Normalization}
    \begin{algorithmic}[1]
        \State \textbf{Initialize} parameters $\theta$ for policy $\pi$ and $\phi$ for critic $V$
        \State \textbf{Set} hyperparameters as specified in Table~\ref{tab:Algorithm_Hyperparameters}, maximum steps $p_{\text{max}}$, and counter $p = 0$
        
        \While {$p \leq p_{\text{max}}$}
            \State \textbf{Initialize} buffer $D = \varnothing$, mask $M = \varnothing$ and $b = 0$
            \State // Interact in $E$ environments, each with $N$ drones
            \While {$b \leq E \times L \times N$} // The batch size
                \For {$t = 1$ \textbf{to} $t_{\text{max}}$}
                    \For {$i=1$ \textbf{to} $N$}
                        \State $\mathbf{o}_i^t = \mathcal{O}(\mathbf{s}^t;i)$ // Receive local observation
                        \State $\mathbf{a}_{i}^{t} \sim \pi_{\theta}(\mathbf{o}_{i}^{t})$ // Sample action
                        \State $\tilde{v}_{i}^{t} = \textsc{Normalize}(V_{\phi}(\mathbf{o}_{i}^{t}); \mu, \sigma)$
                    \EndFor
                    
                    \State Execute joint actions $\mathbf{a}^t$, receive rewards $\mathbf{r}^t$, 
                    \Statex \hspace{4.2em} values $\tilde{\mathbf{v}}^t$ and crash masks $\mathbf{m}^t$
                    \State $D = D \cup \{\mathbf{o}^t, \mathbf{a}^t, \mathbf{r}^t, \tilde{\mathbf{v}}^t\}$
                    \State $M = M \cup \{\mathbf{m}^t\}$ // Store mask for crashes
                    \State $p = p + 1$, $b = b + N$ // Update counters
                \EndFor
            \EndWhile
            
            \State Denormalize value estimates $\tilde{\mathbf{v}}^t$ to obtain $\mathbf{v}^t$
            \State Compute advantages $\hat{A}$, returns $\hat{R}$ using $\mathbf{v}^t$ and 
            \Statex \hspace{1.2em} update value normalization parameters, $\mu$, $\sigma$
            \State $D_{\text{masked}} = D \odot M$ // Mask invalid experiences
            
            \State Update $\theta$ and $\phi$ using PPO loss over $K$ epochs with 
            \Statex \hspace{1.2em} masked data $D_{\text{masked}}$
        \EndWhile
    \end{algorithmic}
    \label{algo:IPPO}
\end{algorithm}

\begin{table}[htbp]
    \vspace{-0.3cm}
    \caption{The Hyperparameters of Our Training Algorithm}
    \begin{center}
        \renewcommand{\arraystretch}{1.2}
        \begin{tabular}{ll|ll}
            \Xhline{1pt}
            \textbf{Parameter} & \textbf{Value} & \textbf{Parameter} & \textbf{Value} \\ \hline
            Discount factor $\gamma$ & $0.99$ & PPO clip range $\varepsilon$ & $0.2$ \\
            GAE parameter $\lambda$ & $0.95$ & Learning rate & $3 \times 10^{\text{-}4}$ \\
            FC layer dimension & $128$ & Number of hidden layers & $2$ \\
            Activation function & tanh & Max episode length $t_{\text{max}}$ & $1500$ \\
            Training epochs $K$ & $10$ & Buffer length $L$ & $4096$ \\ \Xhline{1pt}
        \end{tabular}
    \end{center}
    \label{tab:Algorithm_Hyperparameters}
    \vspace{-0.3cm}
\end{table}

\subsubsection{\textbf{Quadrotor dynamics}} \label{sec:dynamics}
To facilitate efficient large-scale RL training and simplify implementation, we adopt a simplified nonlinear quadrotor dynamic model based on \cite{foehn2021time}, which reduces computational complexity compared to high-fidelity 3D simulations. The dynamics of each vehicle are given by:
\begin{align}
    {\dot{\mathbf{x}}} = 
    \begin{bmatrix}
        {\dot{\mathbf{p}}} \\ {\dot{\mathbf{v}}} \\ {\dot{\mathbf{q}}}
    \end{bmatrix} = 
    \begin{cases}
        {\mathbf{v}} \\
        \mathbf{g}+\mathbf{R}({\mathbf{q}}){{\mathbf{z}_B}}{T} - \mathbf{R}({\mathbf{q}})\mathbf{D}\mathbf{R}^T({\mathbf{q}})\mathbf{v} \\
        \frac{1}{2}\Lambda({\mathbf{q}})
        \left[ 0 \quad {\boldsymbol{\omega}} \right]^T \\
    \end{cases},
    \label{equ:quadrotor_dynamics}
\end{align}
where ${\mathbf{z}_B}$ is the z-axis unit vector in the body frame, ${T}$ is the mass-normalized thrust, and $\mathbf{R}({\mathbf{q}})$ denotes the rotation matrix. The variables ${\mathbf{p}}$, ${\mathbf{v}}$,  ${\mathbf{q}}$ and ${\boldsymbol{\omega}}$ represent the position, inertial velocity, attitude quaternion, and body rate, respectively. Aerodynamic effects are modeled as linear terms \cite{faessler2017differential} of body velocity, which are approximated by the diagonal matrix $\mathbf{D}=-\text{diag}(d_x,d_y,d_z)$.

The network outputs the desired thrust ${T_d}$ and body rates ${\boldsymbol{\omega}}_{d}$, which are sent to a low-level PID controller. Time constants ${k_t}$ and ${k_{\omega}}$ model the delay between the commands and the actual control outputs:
\begin{align}
    \begin{bmatrix}
        {\dot{T}} \\ {\dot{\boldsymbol{\omega}}}
    \end{bmatrix} = 
    \begin{bmatrix}
        {{1/k_t}({T_d} - {T})} \\
        {{1/k_{\omega}}({\boldsymbol{\omega}}_{d} - {\boldsymbol{\omega}})}
    \end{bmatrix}.
\end{align}

\begin{table}[htbp]
    \vspace{-0.2cm}
    \caption{Simulation and Reward Parameters}
    \begin{center}
        \renewcommand{\arraystretch}{1.2}
        \begin{tabular}{ll|ll}
            \Xhline{1pt}
            \textbf{Parameter} & \textbf{Value} & \textbf{Parameter} & \textbf{Value} \\ \hline
            $T_{\text{max}}$ & $3.5$ & $\boldsymbol{\omega}_{\text{max}}$ [rad/s] & $[10, 10, 0.3]$ \\
            $[d_x, d_y, d_z]$ & $[0.29, 0.29, 0.38]$ & $\Delta t$ [s] & $0.01$ \\
            $k_t$ [s] & $0.05$ & $k_\omega$ [s] & $0.05$ \\ 
            $k_p$ [m] & $[16, 16, 3]$ & $k_v$ [m/s] & $[15, 15, 5]$ \\ 
            $k_{rp}$ [m] & $[8, 8, 3]$ & $k_{rv}$ [m/s] & $[15, 15, 5]$ \\ 
            $k_d$ [m] & $4$ & $r_{\text{arrival}}$ & $5$ \\ 
            $r_{\text{collision}}$ & $0.5$ & $r_{\text{truncated}}$ & $30$ \\ 
            $\lambda_1$ & $2 \times 10^{\text{-}4}$ & $\lambda_2$ & $1 \times 10^{\text{-}4}$ \\ 
            $\lambda_3$ & $2.4$ & $\lambda_4$ & $0.5$ \\ \Xhline{1pt}
        \end{tabular}
    \end{center}
    \label{tab:Simulation_Parameters}
    \vspace{-0.2cm}
\end{table}

\section{SIMULATION RESULTS AND ANALYSIS}
This section details the training protocol and evaluates our approach through numerical simulations across diverse race tracks. We benchmark against a single-drone baseline policy that uses a reward design based on \cite{song2023reaching} with identical quadrotor specifications. Despite a slight reduction in performance, the multi-drone approach achieves competitive results with low collision rates, demonstrating its effectiveness in environments requiring highly agile maneuvers and dynamic quadrotor interactions.

\begin{figure*}[htbp]
    \centering
    \includegraphics[width = 1.0\textwidth]{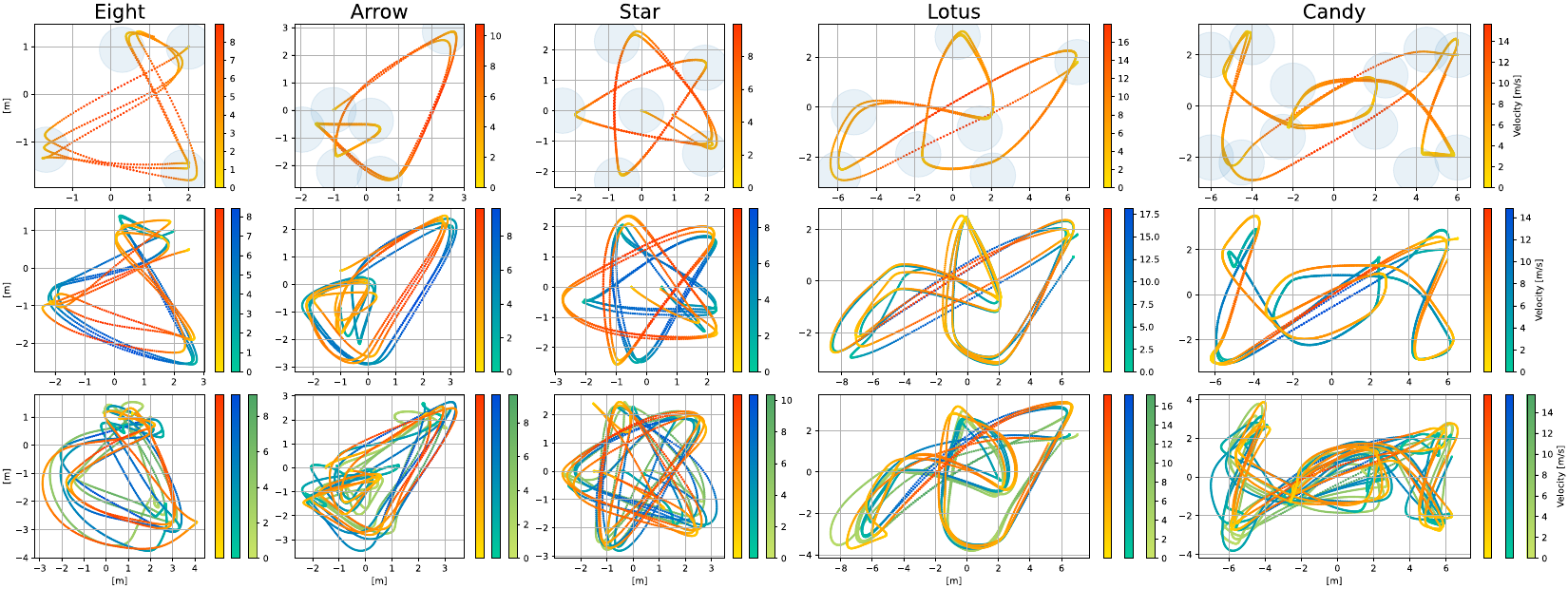} 
    \caption{Visualized quadrotor trajectories on different tracks for single, two, and three quadrotors. Each track has undergone thousands of tests with added noise to the waypoint positions. The plots show quadrotors completing three continuous laps, with shaded areas in the first row marking the waypoints within a radius of $d_{\text{w}}$ for a single lap. Despite a slight reduction in performance, the multi-drone approach achieves competitive results with low collision probability, proving effective across various scenarios.} 
    \label{fig:simulation_plot}
\end{figure*}

\begin{table*}[htbp]
    \caption{Performance Metrics for Different Numbers of Quadrotors Across Five Tracks}
    \begin{center}
        \renewcommand{\arraystretch}{1.2}
        \begin{tabular}{>{\arraybackslash}p{1.0cm}|>{\centering\arraybackslash}p{2.0cm}>{\centering\arraybackslash}p{3.3cm}>{\centering\arraybackslash}p{3.3cm}|>{\centering\arraybackslash}p{0.4cm}>{\centering\arraybackslash}p{0.6cm}>{\centering\arraybackslash}p{0.6cm}|>{\centering\arraybackslash}p{0.6cm}>{\centering\arraybackslash}p{0.6cm}>{\centering\arraybackslash}p{0.6cm}}
            \Xhline{1pt}
            \multirow{2}{*}{\textbf{Track}} & \multicolumn{3}{c|}{\textbf{Lap Time} (s)} & \multicolumn{3}{c|}{\textbf{Collision Rate} (\%)} & \multicolumn{3}{c}{\textbf{Success Rate} (\%)} \\ \cline{2-10}
             & 1 & 2 & 3 & 1 & 2 & 3 & 1 & 2 & 3 \\ \Xhline{1pt}
            Eight & $2.648 \pm 0.010$ & $3.382 \pm 0.010$ & $3.990 \pm 0.134$ & $-$ & $0$ & $10.5$ & $100$ & $100$ & $84.2$ \\
            Arrow & $3.510 \pm 0.010$ & $4.110 \pm 0.060$ & $5.668 \pm 0.585$ & $-$ & $0.2$ & $7.2$ & $100$ & $99.8$ & $89.2$ \\
            Star & $4.104 \pm 0.003$ & $4.627 \pm 0.006$ & $5.376 \pm 0.105$ & $-$ & $0$ & $12.3$ & $100$ & $100$ & $81.6$ \\
            Lotus & $5.295 \pm 0.005$ & $6.514 \pm 0.008$ & $7.033 \pm 0.131$ & $-$ & $0$ & $15.1$ & $100$ & $100$ & $77.2$ \\
            Candy & $8.317 \pm 0.059$ & $10.147 \pm 0.051$ & $11.307 \pm 0.529$ & $-$ & $1.4$ & $24.5$ & $100$ & $98.6$ & $63.2$ \\ \Xhline{1pt}
        \end{tabular}
    \end{center}
    \label{tab:Numerical_Results}
    \vspace{-0.3cm}
\end{table*}

\subsection{Training details}
We train our policy using a modified version of PPO based on \textit{Stable Baselines3} \cite{raffin2021stable} and MAPPO \cite{NEURIPS_MAPPO}. Our learning environment is a heavily customized version of the open-source \textit{gym-pybullet-drones} \cite{panerati2021learning} repository, adapted to our specific requirements.

In each episode, quadrotors are randomly initialized within a fixed cubic volume and required to traverse sequential waypoints with collision avoidance. We employ 72 parallel environments (running at 100Hz with 1500 steps per episode) for experience collection, each containing $N$ agents. Both the policy and value networks are implemented as 2-layer MLPs with 128 units per layer and tanh activation functions. The policy network includes an additional tanh output layer to constrain actions within $[-1,1]$. The simulation and reward details are provided in Table~\ref{tab:Simulation_Parameters}. 

To balance time-optimal performance and safety, we adjust the safe radius based on the number of quadrotors, and the required training time also varies with the number of quadrotors (see Table~\ref{tab:different_Parameters}). All experiments are conducted on a workstation with an AMD Ryzen R9-7950X CPU, NVIDIA RTX 4090 D GPU, and 64 GB DDR5 RAM.

\begin{table}[htbp]
    \vspace{-0.1cm}
    \caption{Safety and Training Efficiency Metrics}
    \begin{center}
        \renewcommand{\arraystretch}{1.2}
        \begin{tabular}{l|cccc}
            \Xhline{1pt}
            \multirow{2}{*}{\textbf{Metric}} & \multicolumn{4}{c}{\textbf{Number of Quadrotors}} \\ \cline{2-5}
             & 1 & 2 & 3 & 5 \\ \hline
            Safe Radius $R$ [m] & $-$ & $0.10$ & $0.16$ & $0.16$ \\
            Convergence Steps & $3 \times 10^7$ & $4 \times 10^7$ & $6 \times 10^7$ & $6 \times 10^7$ \\
            Training Time & 21min & 43min & 1h 20min & 1h 52min \\ \Xhline{1pt}
        \end{tabular}
    \end{center}
    \label{tab:different_Parameters}
\end{table}

\begin{table}[htbp]
    \vspace{-0.5cm}
    \caption{Normalization Parameters for the Five-Drone Configuration}
    \centering
    \renewcommand{\arraystretch}{1.2}
    \begin{tabular}{>{\arraybackslash}p{1.8cm}>
                     {\arraybackslash}p{3.6cm}>{\arraybackslash}p{1.8cm}}
        \Xhline{1pt}
        \textbf{Parameter} & \textbf{Physical Meaning} & \textbf{Value} \\ 
        \Xhline{1pt}
        $k_p$ (m)     & Waypoint position scaling    & $[40, 40, 10]$      \\
        $k_v$ (m/s)   & Ego-velocity scaling         & $[25, 25, 10]$      \\
        $k_{rp}$ (m)  & Relative position scaling    & $[20, 20, 10]$      \\
        $k_{rv}$ (m/s)& Relative velocity scaling    & $[25, 25, 10]$      \\
        $k_d$ (m)     & Distance threshold scaling   & $10$                    \\ 
        $d_{\text{w}}$ [m] & Waypoint radius & $2.5$ \\
        \Xhline{1pt}
    \end{tabular}
    \vspace{-0.3cm}
    \label{tab:5uav_Parameters}
\end{table}

\begin{figure}[htbp]
    \centering
    \includegraphics[width = 0.48\textwidth, trim={200 155 190 160}, clip]{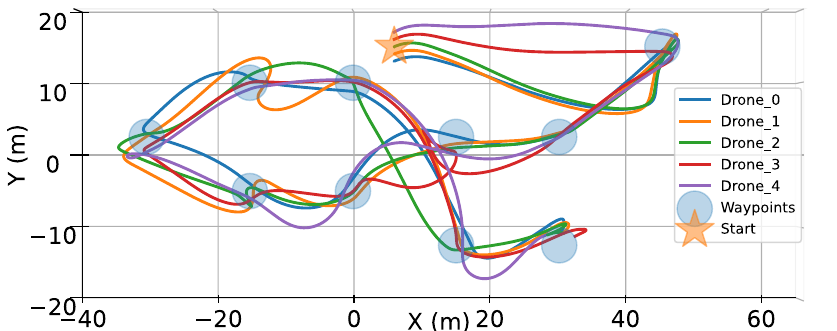} 
    \caption{Trajectories of five quadrotors on the 2019 \textit{AlphaPilot} Challenge course using our decentralized policy, reaching a peak speed of 27.1 m/s, a collision rate of 5.9\%, and a success rate of 83.8\%.}
    \label{fig:plot_for_5uav}
    \vspace{-0.4cm}
\end{figure}

\subsection{Simulation Results}
We validate the policy through thousands of simulation trials on various tracks to evaluate the proposed method, incorporating Gaussian noise into the waypoint positions. We integrate the drones' dynamics at 1000 Hz and set the control frequency to 100 Hz, with a collision threshold of 0.2 m.

\subsubsection{\textbf{Comparative Simulation Experiments}}
Figure~\ref{fig:simulation_plot} shows the comparative experiments of 1, 2, and 3 quadrotors. The key metrics in Table~\ref{tab:Numerical_Results} reveal:
\begin{itemize}
    \item \textit{Success Rate}: Percentage of trials where all quadrotors complete three continuous laps;
    \item \textit{Collision Rate}: Mean collisions per quadrotor per trial.
\end{itemize}
The results show that our approach achieves near-time-optimal flight for multiple drones. Despite slight trade-offs compared to the single-drone policy, the multi-drone policies consistently maintain low collision rates and high success rates, ensuring a balance between time efficiency and safety.

\subsubsection{\textbf{Evaluation of a Five-Drone System}}
To accommodate five quadrotors, we use the 2019 AlphaPilot Challenge \cite{guerra2019flightgoggles} track (Figure~\ref{fig:plot_for_5uav}). With new normalization parameters outlined in Table~\ref{tab:5uav_Parameters}, results show a maximum speed of 27.1 m/s, a collision rate of 5.9\%, and a success rate of 83.8\%.

\section{EXPERIMENT SETUP AND RESULT}
In this section, we conduct real-world flights using a 
305~g quadrotor that can produce a thrust-to-weight ratio of 3.5, as shown in Figure~\ref{fig:real_platform}. The quadrotor runs Betaflight for low-level control and uses a Cool Pi computer for onboard policy inference at 100 Hz via LibTorch. An Optitrack motion capture system provides position and attitude measurements in a $5.5\,\text{m} \times 5.5\,\text{m} \times 2.0\,\text{m}$ space. Notably, we directly deploy the simulation-trained network without fine-tuning, despite unmodeled dynamics and sensor noise.

\subsubsection{\textbf{Star Track Flight}}
\begin{figure}[htbp]
    \centering
    \includegraphics[width = 0.48\textwidth, trim={2.2cm 0.75cm 2.5cm 2cm}, clip]{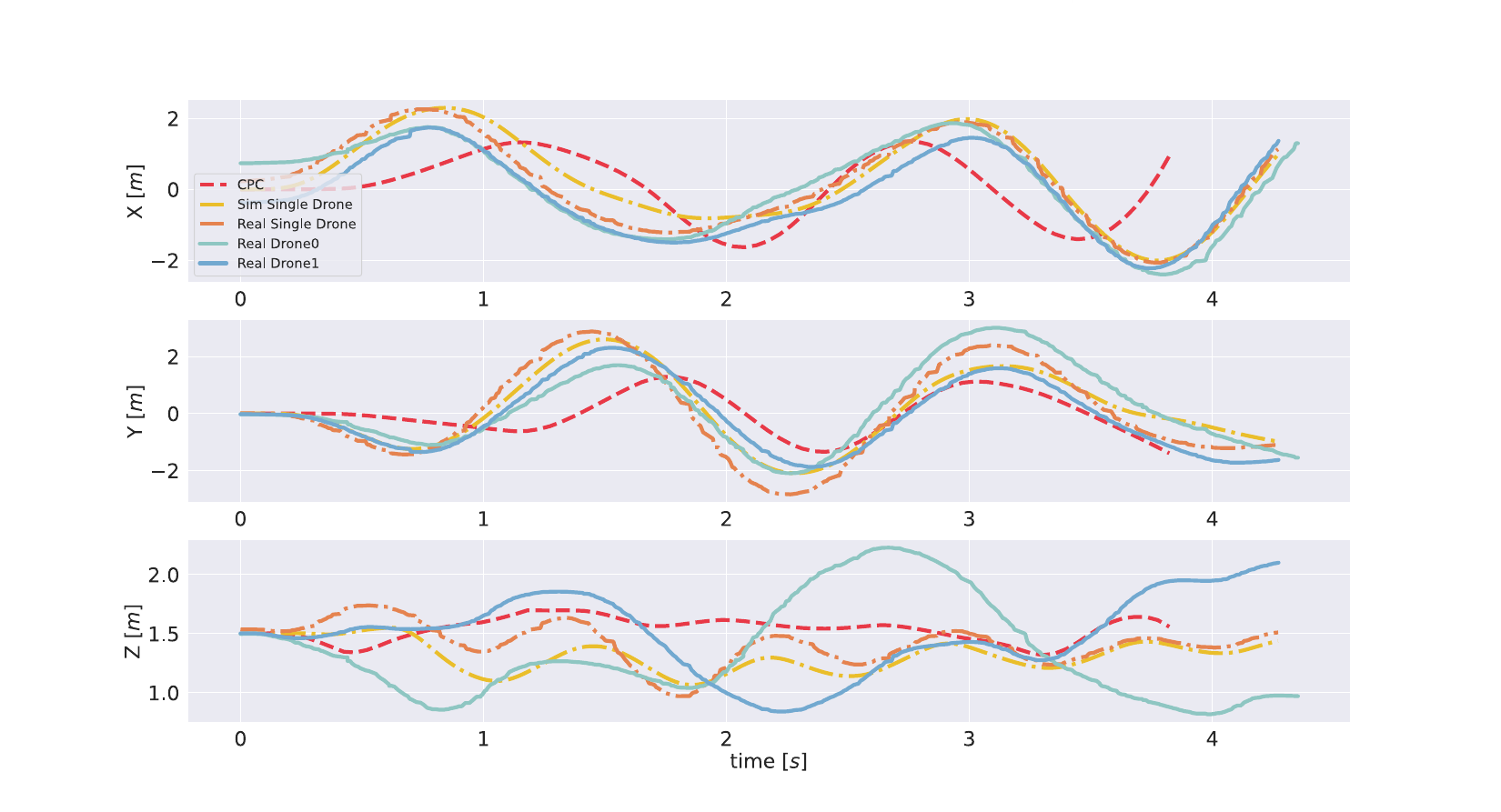} 
    \caption{Time-history of position coordinates during Star track experiments: the time-optimal CPC method (red dashed), the single-drone simulation (yellow dash-dotted), the single-drone real-world flight (orange dash-dotted), the two-drone real flights (solid). All data corresponds to first-lap performance under identical initial conditions.}
    \label{fig:Time-domain_plot}
\end{figure}

To assess the performance of the proposed method, we compared the two-drone policy network with both the CPC method \cite{foehn2021time} and the single-drone policy network on the Star track (Figure~\ref{fig:simulation_plot}). The experiment spans 3 consecutive laps, with flight data from the first lap for different methods shown in Figure~\ref{fig:Time-domain_plot}.

The experimental results indicate that the single-drone real-world performance closely aligns with the simulations, exhibiting only a slight sim-to-real gap of 0.011~s. The single-drone policy is 0.44~s slower compared to the CPC method, highlighting its near-time-optimal performance. The two-drone setup has a mean lag of 0.041~s compared to the single-drone case.

\subsubsection{\textbf{Moving Waypoints and Aggressive Head-on Flight}}
\begin{figure}[hbtp]
    \centering
    \includegraphics[width=0.48\textwidth, trim={1.6cm 0cm 1.6cm 0cm},clip]{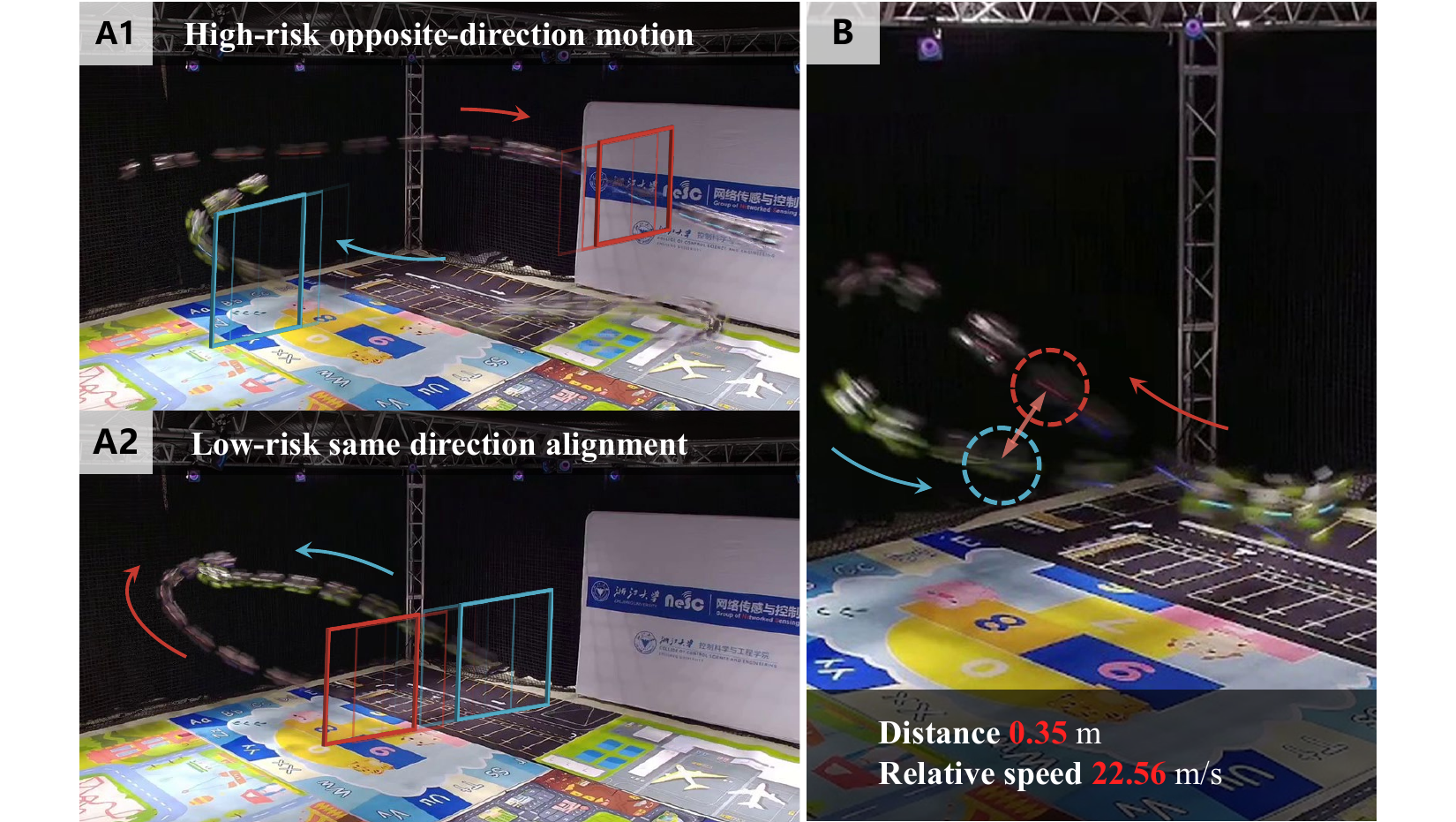}
    \caption{Real-world challenging flight snapshots. (A) High-speed traversal through moving gates featuring millisecond-level replanning and a peak relative speed of 26.07 m/s, (B) aggressive head-on flight maintaining minimum separation distance of 0.35 m. Agents autonomously develop coordination strategies without explicit training.}
    \label{fig:moving}
\end{figure}

Figure~\ref{fig:moving} shows the policy's performance under challenging tasks. Despite extreme dynamic conditions, the two quadrotors maintain a minimum relative distance of 0.35~m and reach a maximum absolute speed of 13.65~m/s (26.07~m/s relative) and a peak body rate of 13.4~rad/s, which highlights the method's capability to realize millisecond-level replanning in highly dynamic environments with agile maneuvers.

Interestingly, the agents exhibit emerging self-organized behaviors—such as (A) adapting from high-risk opposite-direction motion to low-risk same-direction alignment, and (B) employing a near-miss strategy to skillfully avoid collisions rather than abrupt braking—all without the need for global planning or a coordination reward in the training objective (see \href{https://youtu.be/KACuFMtGGpo}{\textbf{\emph{video}}}\footnotemark[3] demonstration for details).

Overall, the proposed network demonstrates strong performance, closely aligning with simulations while effectively handling highly dynamic interactions. The network can be directly transposed from simulation to real-world flights without architectural modifications or fine-tuning.

\section{CONCLUSIONS}

In this study, we train decentralized policy networks for multi-drone time-optimal motion planning. The networks enable drones to navigate through the waypoints with near-time-optimal maneuvers while avoiding collisions, facilitating high-frequency inference onboard.

We design an efficient reward structure with a soft collision-free mechanism, encouraging gradual collision avoidance learning while prioritizing primary tasks. To enhance learning efficiency and stability, we leverage a customized PPO in a \textit{centralized training, decentralized execution} fashion, incorporating invalid-experience masking and value normalization. A simplified quadrotor dynamic model is adapted for ease of implementation. Extensive simulations and real-world experiments demonstrate that the multi-drone policies deliver strong performance in dynamic scenarios with aggressive maneuvers despite minor trade-offs.

Future research could focus on three directions: First, enhancing temporal reasoning through LSTM-based trajectory prediction for dynamic obstacle handling. Second, incorporating LiDAR/event camera observations to enable GPS-denied navigation. Third, developing team reward mechanisms for coordinated tasks like formation flight. These advancements may pave the way for combining aggressive maneuvers with traditional mission tasks, unlocking unprecedented capabilities.

\bibliography{root}
\bibliographystyle{ieeetr}

\end{document}